%% file: paper.tex
\documentclass[a4paper]{esannV2}
\usepackage[T1]{fontenc}
\usepackage{lmodern}
\usepackage[utf8]{inputenc}
\usepackage{graphicx,amsmath,amssymb,array,booktabs,url}
\newcommand{\F}{F_1}

\begin{document}
\title{When Disagreement Labels Do Not Identify\protect\\F1 Rankings}
\author{Roberto Fern\'andez-Barrios, Iker Pastor-L\'opez,\\
Asier Gonz\'alez-Santocildes and Pablo Garc\'ia Bringas
\vspace{.3cm}\\
Faculty of Engineering, University of Deusto\\
Avda. de las Universidades, 24, 48007 Bilbao, Spain
}
\maketitle
\begin{abstract}
Agreement labels cancel from an accuracy comparison but can change an F1
ranking. We specialize partial identification to an explicit audit after
labeling disagreements: two agreement counts give exact bounds, certifying a
ranking or a choice within a specified F1 tolerance under every completion.
Across 3,200 frozen model pairs, shared completions tighten bounds and a 0.01
tolerance reduces labeling costs. A zero-loss (weak-winner) control attributes
most savings to positive tolerance, not ties. Guarantees concern the evaluated
batch, not future data.
\end{abstract}

\section{What can disagreement labels decide?}
Suppose two fixed classifiers disagree on three of ten items. One predicts
positive on all three, the other negative; only one of these labels is positive.
They agree positive on three other items and negative on four. Even knowing
that exactly four of the ten labels are positive does not settle their F1
ranking. Allocating the three remaining positives to negative agreements gives
F1 scores $(1/5,0)$; allocating them to positive agreements gives $(4/5,6/7)$.
The same disagreement labels and prevalence support opposite winners.

Given predictions, disagreement labels and optional prevalence, we reduce
unknown agreement labels to two counts. Their exact F1 bounds certify a choice
at the requested tolerance or require agreement queries, preserving a shared
label completion.

Discordant-pair evaluation by Musgrove et al.\ \cite{musgrove2023} uses
prevalence and baseline sensitivity/specificity assumed to transfer to the
validation data. We have no such baseline-performance information; the example
above therefore does not refute their method. Active F-measure estimation \cite{sawade2010} and Okanovic et al.'s
Bayesian, accuracy-based best-or-near-best selection \cite{okanovic2025}
address label efficiency without our deterministic all-completion F1 guarantee.
F1 batch dependence \cite{lipton2014} and single-model missing-label bounds
\cite{dervovic2025} are established. We give an explicit specialization of
joint identification frameworks \cite{kuboyama2026,tdc,polo2024}.

\section{An explicit audit criterion}
Fix binary predictions $q,c\in\{0,1\}^n$ and correct unknown labels.
Let $a,b$ count disagreements $(q,c)=(1,0),(0,1)$;
$n_{00},n_{11}$ count agreements. After all $D=a+b$ disagreement labels are
revealed, let $u,v$ be their respective positive counts. Unknown positive
counts on $(0,0)$ and $(1,1)$ agreements are $s,t$, respectively. Write
\[
 P=u+v+s+t,\quad Q=a+n_{11},\quad C=b+n_{11},\quad d=u-v.
\]
$P$ is the true positive count, while $Q,C$ are observable predicted counts.
Positive-class F1 is $2\mathrm{TP}/(P+\text{predicted positives})$, with
$0/0=0$. The paired difference is
\begin{equation}\label{eq:gap}
 \Delta(s,t)=\F(q)-\F(c)=\frac{2(t+u)}{P+Q}-\frac{2(t+v)}{P+C}.
\end{equation}
By contrast, accuracy difference $(2d+b-a)/n$ is already determined;
balanced accuracy difference $[nd+(b-a)P]/[2P(n-P)]$ is determined if
$0<P<n$ is known, and undefined at $P=0,n$.

\noindent\textbf{Sharp sign criterion (specialization).}
Set $e=d+b-a$, $K=d(u+v+n_{11})+ub-va$. Then $\Delta$ has the sign of
\begin{equation}\label{eq:affine}
 S(s,t)=K+ds+et.
\end{equation}
Without prevalence information, agreement queries leave an integer rectangle
$s\in[s_-,s_+]$, $t\in[t_-,t_+]$, initially $[0,n_{00}]\times[0,n_{11}]$.
Its exact numerator bounds are
\begin{align}\label{eq:bounds}
 S_-&=K+\min(ds_-,ds_+)+\min(et_-,et_+),\nonumber\\
 S_+&=K+\max(ds_-,ds_+)+\max(et_-,et_+).
\end{align}
Both strict rankings are possible iff $S_-<0<S_+$; $S_->0$ or $S_+<0$
identifies a strict winner, and $S_-=S_+=0$ an exact tie. Other cases allow a tie
and one winner.

\noindent\textit{Proof.} Cross multiplication gives
$\Delta(P+Q)(P+C)/2=dP+(C-Q)t+uC-vQ=K+ds+et$.
The denominator is positive unless $P=0$ and a predictor has no positives;
then both F1 scores and $S$ are zero. Every integer count in the rectangle
is attainable. An affine function reaches its extrema at its corners,
proving the sign claims.\hfill$\square$

\smallskip\noindent\textbf{Equal-count corollary.}
If $Q=C=R>0$, then $\Delta=2d/(P+R)$: disagreement labels identify the sign
without knowing $P$. If $R=0$, both scores vanish. Magnitude can remain unknown.

\smallskip\noindent\textbf{Exact magnitude bounds.}
For fixed $P>0$, the difference is affine in $t$:
\begin{equation}\label{eq:fixed}
 f_P(t)=\frac{2u}{P+Q}-\frac{2v}{P+C}
          +\frac{2(C-Q)t}{(P+Q)(P+C)}.
\end{equation}
The feasible endpoints are
$\ell_P=\max(t_-,P-u-v-s_+)$ and
$h_P=\min(t_+,P-u-v-s_-)$.
If $\ell_P\le h_P$, evaluating $f_P$ at both endpoints gives its exact hull;
otherwise that $P$ is infeasible. At $P=0$ both scores are zero. With unknown
$P$, enumerate feasible integer counts $P$ and take the smallest and largest
of these endpoints, yielding the exact interval hull $[L,U]$ in $O(n)$ time.
This includes $P=n$. The sign proof does not place unknown-$P$
magnitude extrema at rectangle corners.

\smallskip\noindent\textbf{A choice with bounded regret.}
For absolute F1 tolerance $\varepsilon\ge0$, the worst losses from selecting
$q$ and $c$ are $R_q=\max(0,-L)$ and $R_c=\max(0,U)$.
Choose the smaller, breaking ties in favor of $q$, and stop iff
\begin{equation}\label{eq:stop}
 \min(R_q,R_c)\le\varepsilon
 \quad\Longleftrightarrow\quad L\ge-\varepsilon\ \text{or}\ U\le\varepsilon.
\end{equation}
This is Kuboyama's tolerance relaxation and minimum worst-case excess choice
\cite[\S3, \S5]{kuboyama2026} on the F1 hull. These losses prove the batch guarantee.
It does not require $[L,U]\subseteq[-\varepsilon,\varepsilon]$.
At $\varepsilon=0$ it allows a weak winner before distinguishing a strict win
from an exact tie. Failure means $L<-\varepsilon$ and $U>\varepsilon$: opposite compatible
advantages exceeding the tolerance. The inclusive material-reversal condition $L\le-\varepsilon$,
$U\ge\varepsilon$ differs only at the boundaries.

\medskip\noindent\textbf{Relation to general identification frameworks.}
Kuboyama \cite{kuboyama2026}, Theorem~1, Proposition~1,
\S4.5 and \S6.2, gives joint all-completion certificates, agreement relevance
and tolerance for selective prediction. For $H\in\{Q,C\}$ and known $P$ with $P+H>0$, $1-\F=(\mathrm{FP}+\mathrm{FN})/(P+H)$ is an additive
weighted error. Sharp bounds must nevertheless retain $\sum_i y_i=P$;
his unrestricted-label extrema cannot simply be reused under this constraint.
With unknown $P$, denominators vary across completions. Our contribution is the explicit F1 audit, including degeneracies,
not general certificate, joint-comparison or tolerance principles.

TDC \cite{tdc}, Theorem~1, gives sharp certificates for fixed additive-loss
contrasts over unrestricted completions. With $P>0$ fixed,
\begin{equation}\label{eq:additive}
 \Delta=\sum_i y_i\left[\frac{2q_i}{P+Q}-\frac{2c_i}{P+C}\right].
\end{equation}
Imposing $\sum_i y_i=P$ is essential for sharpness; TDC's \S4.5 already groups items by
prediction signature, constrains their positive counts and enumerates prevalence.
Equations (\ref{eq:affine})--(\ref{eq:fixed}) carry this reduction to F1.

Polo et al.\ \cite{polo2024}, Eq.~(1.1) and \S3.2, also cover fixed-$P$ F1:
use uniform batch indices, distinguish audited items and fix the label marginal.
For (\ref{eq:additive}), the remaining fractional labels have an integer sum.
Every vertex is integral: two fractional entries permit a sum-preserving
perturbation; one cannot sum to an integer. Hence linear extrema equal our
completion extrema. Our contribution is the explicit audit and its consequences.

Separate bounds form
$[\min\F(q)-\max\F(c),\allowbreak\max\F(q)-\min\F(c)]$. The feasible set is shared, but the optimizers need not agree.
Marginals at fixed $P$ are affine in $t$, so enumeration is exact.

\section{Retrospective evaluation}
\noindent\textbf{Frozen cases and units.} We reuse TDC's prediction artifact,
release 1.1.6 \cite{tdcartifact}, without refitting models. Each of 36 cells
contains 500 targets, a validation-selected quantum-kernel predictor, and
115 classical configurations. Twenty cells cover TableShift's \cite{tableshift} BRFSS diabetes and ACS
FoodStamps (five seeds, SVC and Gaussian-process classification);
16 cover eight EMBER \cite{ember}, network ToN-IoT \cite{toniot}
and UNSW-NB15 \cite{unsw} scenarios with both classifiers.
ToN-IoT uses the original \texttt{train\_test\_network.csv} scanning-task
variant, not sensor telemetry or a NetFlow re-export. We collect no new data.
Deduplicating identical classical predictions within cells yields 3,200 pairs.
The criterion applies to arbitrary fixed binary predictions.

The pilot (eight cells, 801 pairs) preceded label inspection in this study;
the reserved 28 cells (2,399 pairs) are an analytical holdout of existing evidence.
Tolerance, separate-bound and validation analyses are retrospective extensions.

Fractions average equally across cells; costs average orders, pairs within
cells, then eligible cells. Shared targets and overlapping subsamples preclude
independent-observation inference; summaries are descriptive. Known $P$ is
an exact information oracle. Initial intervals retain $0<P<n$, the original
balanced-accuracy domain; query cohorts and stops allow $P=0,n$.

\smallskip\noindent\textbf{Acquisition and comparators.} We retain three agreement orders:
uniform; alternating shuffled agreement strata; and decreasing $|d|$ or $|e|$
(magnitude, shuffled ties). Seeds 0--31 share permutations across policies and
objectives. Orders use predictions and disagreement labels only; queries shrink
the count intervals. Magnitude applies Kuboyama's static range order \cite[Eq.~(10)]{kuboyama2026}
to $S$: it minimizes numerator width, not necessarily F1 query cost.
We compare strict-sign/exact-tie identification, a weak winner ($\varepsilon=0$),
and $\varepsilon=0.01$ on the same trajectories. Full labels verify regret;
they never select pairs, queries or stopping decisions.

We select single classical comparators by minimum Hamming distance, with
original-order ties, or maximum validation balanced accuracy, with ties broken
lexicographically by configuration.
All 115 validation scores were available; selection preceded the new target
analysis and follows TDC's objective, not F1. Hamming proximity is a sensitivity
analysis, not a deployment rule.

\smallskip\noindent\textbf{Mechanism.} At known $P>0$, interval width is exactly
$2|C-Q|(h_P-\ell_P)/[(P+Q)(P+C)]$: predicted-count imbalance and freedom to
allocate positives between agreement groups jointly drive ambiguity.
Without $P$, initial numerator width is $|d|n_{00}+|e|n_{11}$. All 58 equal-count pairs identify the sign but retain nonzero mean magnitude
uncertainty; disagreement count alone does not explain ambiguity.

\smallskip\noindent\textbf{Initial decisions.} In the 28 reserved
cells, the mean material-reversal fraction is 20.54\% without $P$, and 10.52\%
with exact $P$. These are compatible opposite advantages of at least 0.01, not observed errors
or universal frequencies. No pair lies on the 0.01 boundaries, so complements
are certifiable at that tolerance.

Keeping dependence matters: in the reserved cells, paired bounds
certify 79.46\% of choices at $\varepsilon=0.01$, versus 6.57\% for separate
bounds; with exact $P$, 89.48\% versus 18.80\%. Mean widths are respectively
0.173 versus 1.585, and 0.034 versus 1.005. This is an information ablation
on identical feasible sets, not superiority over joint Kuboyama, TDC or Polo certificates.

\begin{figure}[!ht]
\centering
\includegraphics[width=112mm]{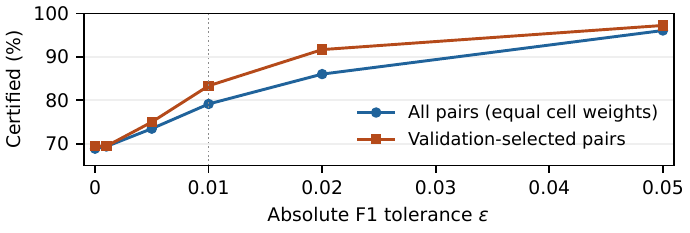}
\caption{Certification after disagreement labeling, unknown prevalence
($0<P<n$). Equal-cell means over all 36 cells: 3,200 pairs, or one
validation-selected pair per cell. Points are evaluated tolerances; lines guide
the eye. Descriptive fractions, not confidence estimates.}
\label{fig:tolerance}
\end{figure}

Figure~\ref{fig:tolerance} shows the tolerance dependence: at 0.01, certification
is 79.15\% over all 36 cells, versus 79.46\% in the 28 reserved cells.
Hamming selection leaves material
ambiguity in 3/36 cells; validation selection leaves
6/36 (both have 11/36 strict reversals). These one-pair-per-cell
designs answer a different question from the all-configuration pool.

\smallskip\noindent\textbf{Label costs.} Table~\ref{tab:costs} includes already-decided pairs. Under magnitude ordering,
a 0.01 tolerance reduces total labels from 121.16 to 74.59 (all pairs), and 92.57 to 40.29
(validation), versus 500 for full labeling. Zero-loss totals, 120.92 and 92.25,
remain close to strict/exact costs: positive tolerance accounts for most savings.

\begin{table}[!ht]
\centering
\setlength{\tabcolsep}{4.5pt}
\begin{tabular}{llrrrrr}
\toprule
 & & Initial & \multicolumn{2}{c}{Additional} & \multicolumn{2}{c}{Total}\\
Cohort & Policy & $D$ & Exact & $\varepsilon$ & Exact & $\varepsilon$\\
\midrule
\input{cost_rows_v4.tex}\\
\bottomrule
\end{tabular}
\caption{Mean labels on common cohorts, 36 cells, 32 orders per policy.
All pairs: 3,200 comparisons; validation: one per cell.
Additional labels query agreements; $\varepsilon=0.01$. Each batch has 500
items. Totals are calculated before rounding.}
\label{tab:costs}
\end{table}

Costs vary across cells: under magnitude ordering and $\varepsilon=0.01$,
the all-pair cell means have minimum/median/maximum 14.0/77.0/188.4 labels.
This is descriptive dispersion, not an uncertainty interval. The overall mean
is 74.59 with equal cell weights, versus 82.05 with equal pair weights.
In 15,466 of 307,200 dependent replays, the certified choice is the realized-worse
model; maximum regret is 0.00927, within tolerance. The guarantee is (\ref{eq:stop}); these are retrospective checks.

\section{Use and limitations}
Given a tolerance, an auditor computes $[L,U]$ from the current count bounds,
then chooses by (\ref{eq:stop}) or queries agreements and updates those bounds.

This specialization assumes fixed predictions and correct labels; its guarantee
does not extend to populations, noisy labels, macro-F1, multiclass or families.
Compatible extremes need not be probable; benchmark reuse limits breadth.
Disagreement-first labeling is imposed. Costs are neither optimal-acquisition
lower bounds nor additive across pairs. Statistical alternatives can trade
assumptions or error probability for fewer labels.
\par\smallskip\noindent\textbf{Availability.} Code, frozen inputs and results (v1.4.0):
\url{https://doi.org/10.5281/zenodo.23647323}.

\smallskip\noindent\textbf{Declarations.}
R.F.-B. conducted the research and writing; coauthors provided scientific
guidance. No specific funding. OpenAI Codex assisted literature, design,
mathematical checks, code, analysis and writing; Claude assisted independent
review. All authors approved the content and accept responsibility.
\begin{footnotesize}

\end{footnotesize}
\end{document}

%% file: cost_rows_v4.tex
All pairs & Uniform & 54.5 & 88.21 & 35.58 & 142.70 & 90.07 \\
All pairs & Alternating & 54.5 & 80.64 & 31.29 & 135.13 & 85.78 \\
All pairs & Magnitude & 54.5 & 66.67 & 20.10 & 121.16 & 74.59 \\
Validation & Uniform & 30.7 & 89.07 & 25.56 & 119.79 & 56.28 \\
Validation & Alternating & 30.7 & 77.58 & 16.34 & 108.30 & 47.07 \\
Validation & Magnitude & 30.7 & 61.84 & 9.57 & 92.57 & 40.29

%% file: paper.bbl
\begin{thebibliography}{99}
\bibitem{musgrove2023} D. Musgrove, A. Radtke, and T. Haddad.
Discordant pair analysis for sample efficient model evaluation.
\emph{Scientific Reports}, 13:20987, 2023.
\bibitem{sawade2010} C. Sawade, N. Landwehr, and T. Scheffer.
Active Estimation of F-Measures. \emph{NeurIPS}, 23, 2010.
\bibitem{okanovic2025} P. Okanovic, A. Kirsch, J. Kasper, T. Hoefler,
A. Krause, and N. M. G\"urel.
All models are wrong, some are useful: Model Selection with Limited Labels.
\emph{AISTATS}, PMLR 258:2035--2043, 2025.
\bibitem{lipton2014} Z. C. Lipton, C. Elkan, and B. Narayanaswamy.
Optimal Thresholding of Classifiers to Maximize F1 Measure.
\emph{ECML PKDD}, 225--239, 2014. doi:10.1007/978-3-662-44851-9\_15.
\bibitem{dervovic2025} D. Dervovic and M. Cashmore.
Model Evaluation in the Dark: Robust Classifier Metrics with Missing Labels.
\emph{AISTATS}, PMLR 258:1909--1917, 2025.
\bibitem{kuboyama2026} T. Kuboyama. How Many Labels Does Model Choice Need?
Certificates and Budgets for Selective Prediction. arXiv:2609.18622v1, 2026.
\bibitem{tdc} R. Fern\'andez-Barrios, I. Pastor-L\'opez, A. Gonz\'alez-Santocildes,
and P. Garc\'ia Bringas. Sharp Target-Domain Certificates for Quantum-Kernel
Advantage under Distribution Shift. \emph{arXiv:2610.00080v1}, 2026.

\bibitem{polo2024} F. M. Polo, S. Maity, M. Yurochkin, M. Banerjee, and Y. Sun.
Weak Supervision Performance Evaluation via Partial Identification.
\emph{NeurIPS}, 37, 2024. arXiv:2312.04601.
\bibitem{tdcartifact} R. Fern\'andez-Barrios et al.
TDC prediction artifact, v1.1.6. Zenodo, 2026.
\url{https://doi.org/10.5281/zenodo.21776862}.
\bibitem{tableshift} J. Gardner, Z. Popovi\'c, and L. Schmidt.
Benchmarking Distribution Shift in Tabular Data with TableShift.
\emph{NeurIPS}, 36, 2023.
\bibitem{ember} H. S. Anderson and P. Roth. EMBER: An Open Dataset for Training
Static PE Malware Machine Learning Models. arXiv:1804.04637v2, 2018.
\bibitem{toniot} N. Moustafa. A new distributed architecture for evaluating
AI-based security systems at the edge: Network TON\_IoT datasets.
\emph{Sustainable Cities and Society}, 72:102994, 2021.
\bibitem{unsw} N. Moustafa and J. Slay. UNSW-NB15: a comprehensive data set
for network intrusion detection systems (UNSW-NB15 network data set).
\emph{MilCIS}, 1--6, 2015.
\end{thebibliography}
